\documentclass[letterpaper, 10 pt, conference]{ieeeconf}  

\usepackage[left=1.91cm,right=1.91cm,top=2.54cm,bottom=2.54cm]{geometry}
\usepackage{booktabs}
\usepackage[T1]{fontenc}
\DeclareTextSymbolDefault{\dh}{T1}
\usepackage{hyperref}
\usepackage{graphicx}
\usepackage{balance}
\usepackage{multirow}
\usepackage{subcaption}

\newcommand{\todo}[1]{\textcolor{red}{TODO: #1}}

\newcommand{\gh}[1]{\textcolor{blue}{G.H: #1}}

\renewcommand{\todo}[1]{}
\renewcommand{\gh}[1]{}
\IEEEoverridecommandlockouts                              

\usepackage{xcolor}

\title{\LARGE \bf
Design and Implementation of Smart Infrastructures and Connected Vehicles in A Mini-city Platform
}
\author{Daniel Vargas$^{2}$, Ethan Haque$^{1}$, Matthew Carroll$^{1}$, 
Daniel Perez$^{2}$,\\ Tyler Roman$^{1}$, Phong Nguyen$^{2}$,
Golnaz Habibi$^{1}$%
\thanks{$^{1}$ University of Oklahoma, 110 W Boyd Street, Norman, OK, USA}%
\thanks{$^{2}$ School of Aerospace and Mechanical Engineering,
        University of Oklahoma, 865 Asp Avenue, Norman, OK,
         USA}%
}

\begin{document}

\maketitle
\thispagestyle{empty}
\pagestyle{empty}

\begin{abstract}
This paper presents a 1/10th scale mini-city platform used as a testing bed for evaluating autonomous and connected vehicles. Using the mini-city platform,  we can evaluate different driving scenarios including human-driven and autonomous driving. We provide a unique, visual feature-rich environment for evaluating computer vision methods. The conducted experiments utilize onboard sensors mounted on a robotic platform we built, allowing them to navigate in a controlled real-world urban environment. The designed city is occupied by cars, stop signs, a variety of residential and business buildings, and complex intersections mimicking an urban area. Furthermore, We have designed an intelligent infrastructure at one of the intersections in the city which helps safer and more efficient navigation in the presence of multiple cars and pedestrians. We have used the mini-city platform for the analysis of three different applications: city mapping, depth estimation in challenging occluded environments, and smart infrastructure for connected vehicles. Our smart infrastructure is among the first to develop and evaluate Vehicle-to-Infrastructure (V2I) communication at intersections. The intersection-related result shows how inaccuracy in perception, including mapping and localization, can affect safety. The proposed mini-city platform can be considered as a baseline environment for developing research and education in intelligent transportation systems. 
\end{abstract}
\section{INTRODUCTION}
As technology and urban mobility evolve, Autonomous Vehicles (AVs) and Intelligent Transportation Systems (ITS) are key factors in shaping the future of urban landscapes. The development of these technologies calls for rigorous and varied testing environments that can ensure safety and functionality. Traditionally, testing AVs algorithms in real-world scenarios presented significant costs as well as safety and practical challenges; therefore, researchers are now turning to simulated environments, such as CARLA \cite{carla} and Unity \cite{hossain2019caias}, which provides controlled settings and great step in regards to simulation testbeds; however, these environments still lack the fidelity and complexities of real-world interactions.
\begin{figure}[h]
    \centering
\includegraphics[width=0.9\columnwidth,trim={0 0 0 0},clip]{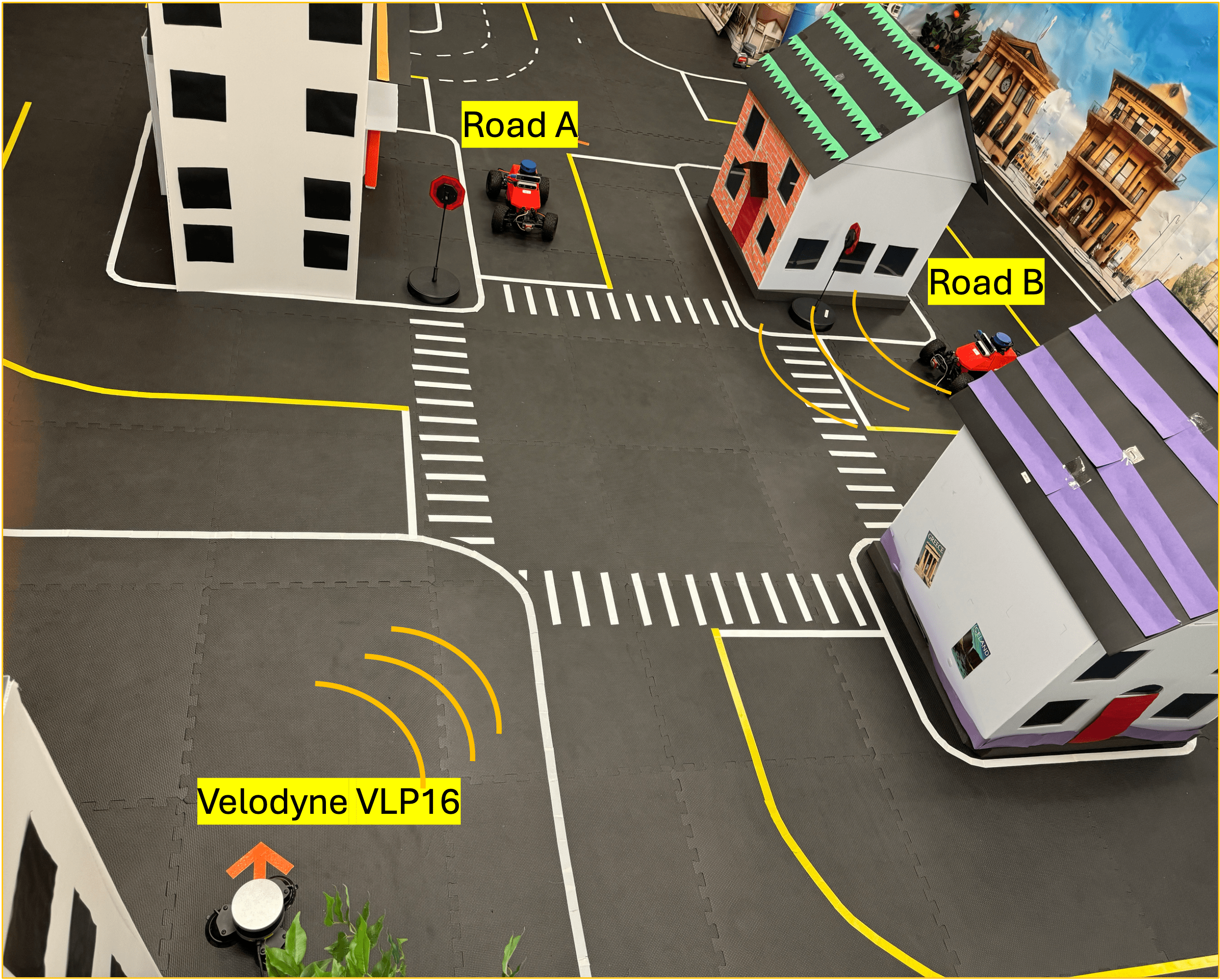}
    \caption{Smart Intersection Scenario: The non-communicating car on road A continues forward without communicating with the infrastructure. communicating car on road B shares its location with the smart infrastructure located at the bottom left corner. The infrastructure warns the communicating car to stop if there is a risk of accident.}
    \label{fig:intersection_scenarios}
\end{figure}
Although advancements have been made in such simulated environments, transitioning these technologies from virtual to complex real-world urban settings remains with interesting challenges. The discrepancy between simulated and real-world conditions can lead to significant performance deviations. To address this gap, scaled testbeds such as mini-cities have provided a more realistic and controllable environment for thorough testing before real-world deployment. A recent comprehensive survey of small-scale testbeds highlights their value in providing cost-effective, controlled environments that bridge the gap between full-scale experiments and simulations~\cite{mokhtarian2024survey}. These testbeds simulate real-world scenarios with varying degrees of complexity and realism, allowing researchers to evaluate algorithms under different conditions. However, these models often lack comprehensive integration of complex urban infrastructures and detailed environmental contexts, which are critical for the next generation of urban mobility solutions.

Prior research has explored other mini-cities for autonomous vehicle testing\cite{buckman2023infrastructure, buckman2022evaluating}, however, these models have yet to integrate key components of ITS such as smart infrastructures and connected vehicles. Our work introduces a novel 1/10th scale mini-city that incorporates these elements, offering a unique test bed that bridges the before-mentioned gap even further between simulation and real-world application. This mini-city not only facilitates complex testing scenarios involving AVs and connected vehicles but also enhances and opens the doors to more research capabilities across various domains, which potentially can accelerate the advancements being made within the fields of autonomous technology and smart infrastructures.

Our proposed mini-city encompasses multiple urban features such as roads, buildings, and traffic systems, as well as a fully developed environmental background. This latter feature surrounds the entire perimeter of the test bed, are scaled appropriately down to the same scale as the rest of the physical components of the mini-city. Adding these layers of environmental context enhances the simulation’s realism and complexity, which had not previously been implemented in other mini-cities, which further makes for a more comprehensive testing and development site for AVs and ITS. 
\subsection{Main contribution}
The main contribution is listed as follows:
\begin{itemize}
    \item Design and build a physical 1/10th scale city including a variety of buildings, mimicking a realistic urban areas.
    \item Design and implement a smart intersection that enables cars to communicate with the infrastructure at the intersection, leading to a safe navigation.
    \item Consider mini-city as a test bed for evaluating autonomous driving-related research such as depth estimation, mapping, and autonomous navigation.
\end{itemize}
\section{Related Work}
Related work can be grouped into small cities, robotic cars, and smart infrastructure.
\subsection{Small City Platform}
There are several studies in mini-city. 
Buckman et al.~\cite{buckman2022evaluating}, presented a 1/10th scale mini-city, which has been used for object detection and state estimation. The work ~\cite{buckman2023infrastructure} presented a mini-city in which the driving behaviors is analyzed to identify vehicle failure. This indicates the importance of vehicle failure identification such as swerving due to manual control, lane offset, speeding, and periodic steering and speeding. However, this mini-city does not utilize connected vehicle communication as our version does.

Duckietown~\cite{Duckietown} is another example of mini-city. The primary goal of this city was to provide high schools and other universities with a pre-assembled bundle to deploy and begin their work on autonomous navigation. Duckietown has a total area of 240 sqft, and includes three traffic lights and thirty duckiebots. The mini-city is two times larger while it is much cheaper in comparison with Duckietown, as shown in \ref{tab:minicity BOM},  which makes our platform affordable to broader community.
\subsection{Robotic Platforms}
There are a variety of robotic platforms with different scales have been used for conducting research in autonomous cars. This section lists only platforms that are closely related to our cars. 
F1TENTH~\cite{F1TENTH} is a race car developed by a team with the same name, it has been used for educational and research purposes. Although this platform shares similar processor and sensors with our robotic platform, our design is based on MuSHR's platform~\cite{srinivasa2019mushr}, which was originally inspired by MIT's RACECAR \cite{7910242}. 
Furthermore, we utilize our cars to conduct research in connected vehicles and smart infrastructure in smart cities.
We have built a fleet of 1/10 scale race car, called ARC. ARC are used for a variety of experiments conducted in mini-city. The cars are equipped with Wi-Fi, which enables them to communicate within themselves as well as smart infrastructure. 
\subsection{Smart Infrastructure and Smart Intersections}
Recent innovations in smart roads, crucial for smart cities, are detailed by Toh et al~\cite{toh2020advances}. The paper discusses the integration of advanced transportation technologies like V2X communication (\emph{i.e.}, Vehicle to everything communication), smart intersections, and automated emergency services. While these advancements significantly enhance road safety and efficiency, the authors note a gap in the implementation of these technologies at a granular, interconnected level across urban infrastructures. Our research aims to address this gap by proposing an integrated framework that enhances vehicle-to-infrastructure communication at smart intersections, particularly in miniaturized urban environments like our mini-city.

\section{Design Mini-city and its Components}
\subsection{Mini-city Layout}
\begin{figure}[]
    \centering
\includegraphics[width=0.9\columnwidth,trim={0 0 0 0},clip]{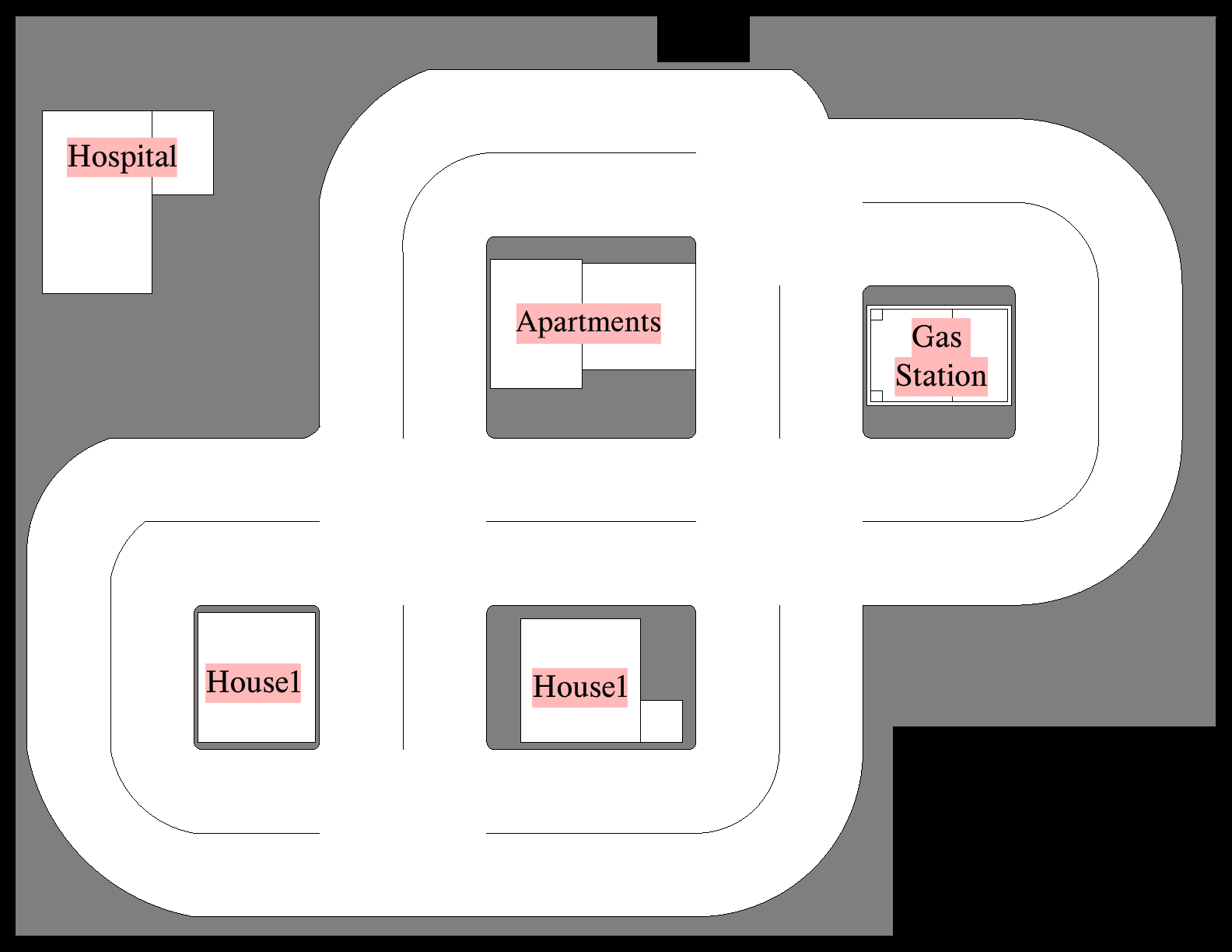}
    \caption{Ground Truth (GT) map of the mini-city}
    \label{fig:Ground Truth}
\end{figure}

\begin{figure*}[]
    \centering
    \begin{subfigure}[]{0.45\textwidth}
     \includegraphics[width=\textwidth, trim={0 320 0 0},clip]{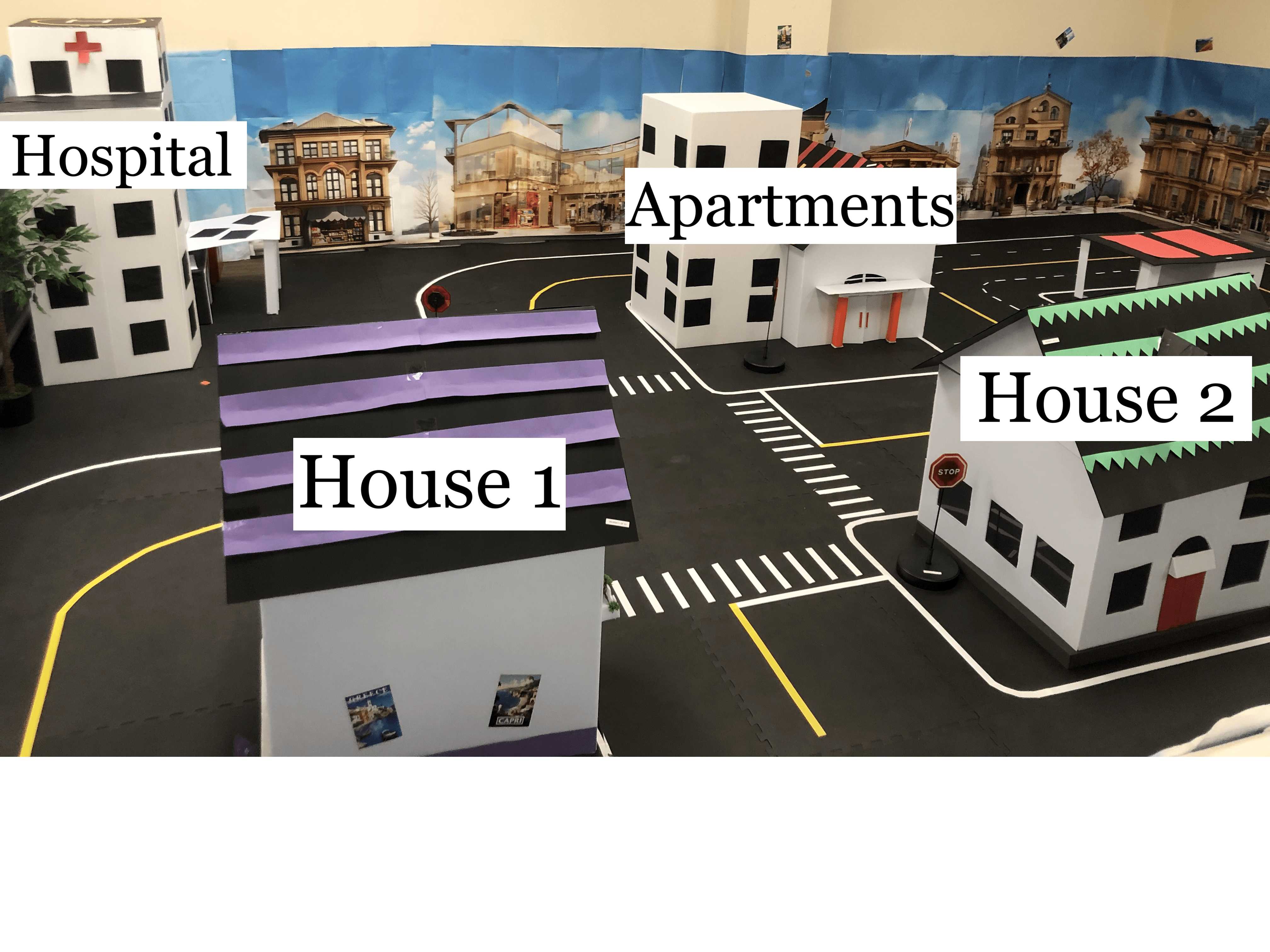}
        \caption{}
    \end{subfigure}%
    \begin{subfigure}[]{0.45\textwidth}
\includegraphics[width=\textwidth, trim={0 0 0 0},clip]{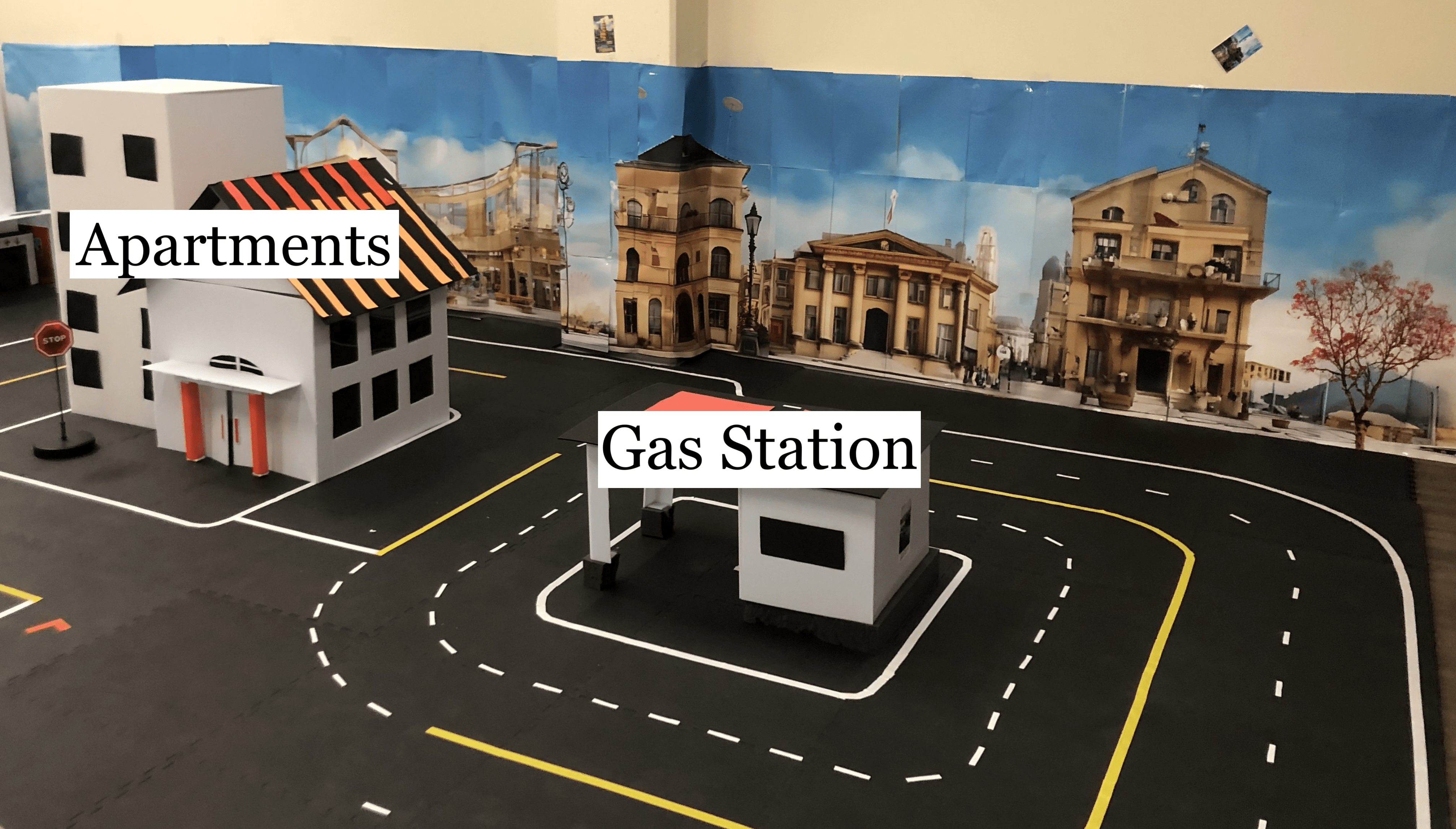}
        \caption{}
    \end{subfigure}
    \caption{(a) West side, and (b) East side of mini-city with labeled buildings. The city background is generated by Artificial Intelligence.}
    \label{fig:minicity}
\end{figure*}

We consider a variety of buildings with different heights,  two-way roads, intersections with realistic layout, and other urban features in general to design and build our mini-city. The proposed mini-city, consists of six size-varied buildings including two houses, one apartment, one hospital, one gas station, and a series of sophisticated traffic infrastructure to resemble driving scenarios in urban areas.  The overall size of the mini-city is $26.75'\times19.75'$ which is around 528 sqft, and it consists of two four-way intersections, two three-way intersections, and two blind-curve areas. The buildings are not attached to the flooring, which makes our design flexible and easily modifiable for different scenarios. Fig.~\ref{fig:Ground Truth} shows the layout of the Mini-city. Fig.~\ref{fig:minicity} shows a detailed illustration of the buildings within the mini-city.
\subsection{Building Mini-city Platform}
We have selected cost-efficient materials to build the mini-city. The overall cost is around \$$1000$.  The bill of the materials with their price is summarized in Table \ref{tab:minicity BOM}. 
We plan to release detailed instructions for building the mini-city. The construction of the city is not only affordable, but also offers a unique layout that provides flexibility for expansion as all of its components, including the buildings and the ground floor, are easily portable. 
All the components of the mini-city is built in 1/10th scale. Roads are made of interlocking 1/2-inch thick foam mats and industrial-grade white and yellow duct tape. The differing colors are for the vehicles to easily differentiate between an outside lane and a median lane. Our team constructed the houses and other buildings from corrugated plastic and black paper to resemble windows. We surrounded the entire city with a feature-rich background and a sky backdrop to assist vehicles with visual navigation in recognizing their location.
\section{Vehicles and Smart infrastructure platforms}
To mimic a city environment, we consider three different platforms: ground vehicles, pedestrians, and ambulance drones. The vehicles were designed and built by members of our team. We considered off-the-shelves humanoid robots to mimic pedestrians\cite{robot}. We also designed and implemented a smart infrastructure to represent a smart city and provide a smart intersection.
\begin{table}[]
    \centering
    \caption{Mini-city Build of Material list}
    \begin{tabular}{|p{1.4cm}|p{2.4cm}|p{0.8cm}|p{1cm}|p{1cm}|} 
    \hline
 Material & Purpose &Quantity & Cost per unit $(\$)$ & Total cost$(\$)$ \\
\hline
 Floor foam & Roads/ground& 4 & 102.23 & 411.72 \\ 
 \hline
 Boards & Buildings& 6 & 39.99 & 239.94 \\
 \hline
 Tape & Road lanes&  3 & 13.28 &  29.84 \\
     \hline
 Fence & Outer wall& 2 & 23.99 & 47.98 \\
     \hline
 PVC Pipes &City enclosure& 16 & 4.71 & 75.36 \\
     \hline
   Posters& Visual Background& 250 & 0.7 & 175\\
    \hline
   Tools/Misc &Assembly/Accessories& 1& 100&100\\
    \hline
 Mini-city &Testing bed&1& 1079.84 &1079.84\\
    \hline
\end{tabular}
    \label{tab:minicity BOM}
\end{table}
\subsection{AirOU RaceCar (ARC)}
\begin{figure}[]
    \centering    
\includegraphics[width=0.4\textwidth, trim={0 100 0 300},clip]{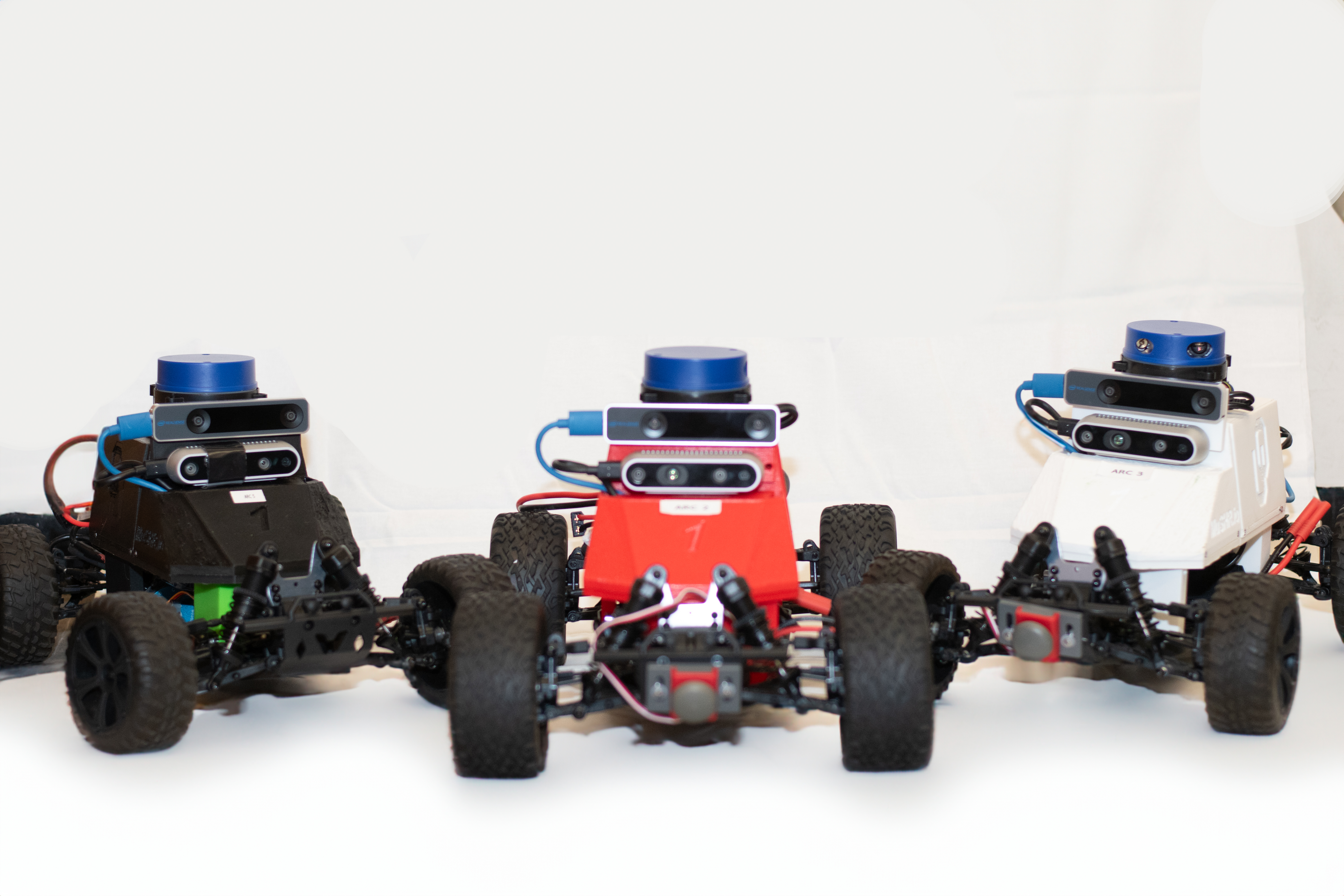}
    \caption{ARC fleet used in the experiments.}
    \label{fig:arc_pic}
\end{figure}

ARCs are 1/10th scale non-holonomic vehicles designed based on MuSHR~\cite{srinivasa2019mushr} that utilizes two cameras, one 2D-LiDAR, and a built-in IMU which are crucial for real-time obstacle detection, simultaneous localization and mapping (SLAM), and path planning. The car dimensions are  $0.51 (W) \times 0.30(L) \times 0.25(H)$, the values are in meters and they weigh approximately $4.5 kg$ with a turning radius of $1.47$ meters. The ARC utilizes an onboard NVIDIA Jetson Nano with 4GB RAM and 128 Cores. The vehicle is controlled via a Flipsky 50A FSESC, which controls the 3500 kv brush-less motor, and a single $20kg$ servo motor that allows us to control the vehicle via Ackermann steering.  Fig.~\ref{fig:arc_pic}
shows our ARC platform. We have modified the original body design in MuSHR and upgraded some of the hardware. This involves adding a tracking camera Intel T265 to enable vision-based localization and mapping. We also considered an $11.1V$ LiPO battery, as the previous version did not provide sufficient current. The previous $7.2V$ LiPO batteries forced us to frequently encounter random shutdowns and issues with starting up in general. After experimenting with a multi-meter, we found the bootup process would demand a large amount of current and since there was an insufficient current, the Jetson Nano would shut off.

\begin{figure}[]
    \centering    \includegraphics[width=.4\textwidth]{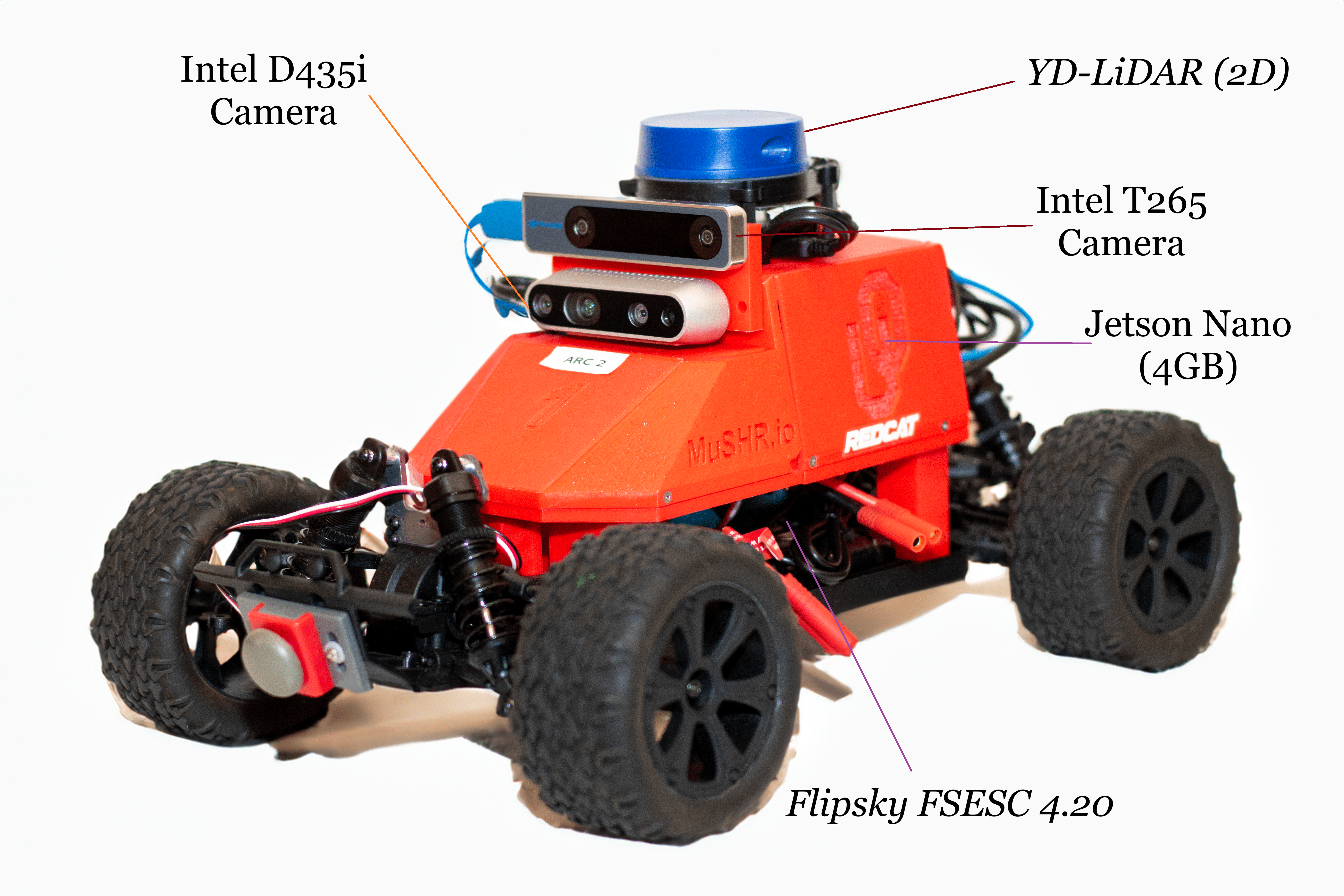}
    \caption{ARC with detailed view of onboard sensors and processor.}
    \label{fig:sensors}
\end{figure}

Fig.~\ref{fig:sensors} shows the ARC components. We have used YDLIDAR X4 2D LiDAR for SLAM, an Intel T265 tracking camera, and an Intel D435i depth camera. All Sensors, motors, and drivers are managed via the onboard computer utilizing Robotics Operating Systems (ROS) to communicate information synced. ROS enables autonomous vehicles and smart infrastructure to share information throughout the city. In this sense, the mini-city emulates a high-tech urban environment where data can be shared between smart devices. 

The car's operating system is Ubuntu 20.04 LTS. We extended MuSHR's software package to support communication between cars and the infrastructure using Wi-Fi. We also extended the computer vision capability of the robots to estimate the depth of different objects in the city environment as well as providing the map using both cameras and LiDAR.

\subsection{Smart Infrastructure}
We designated one intersection in the city to perform our experiments with a smart infrastructure. This platform enables running various experiments in smart-intersection and evaluation of communication between the vehicles and the infrastructure.  
The smart infrastructure, located at the corner of the intersection next to the hospital, is equipped with a 3-D LiDAR (Velodyne VLP-16) which is connected to an NVIDIA Orin Processor. This oversees passing vehicles and pedestrians, sharing real-time data and information regarding the state of the intersection.

Such an installation allows for dynamic traffic control at resolutions beyond conventional induction loop sensors, which only sense the presence of large vehicles in one location at intersections. Such systems are flawed in detecting motorcycles, cyclists, or pedestrians. Our proposed system allows for tracking of different objects, such as cyclists and pedestrians using the 3D point cloud extracted from 3D-LiDAR at intersection, allowing for high-speed tracking and planning at intersections. The implemented system is also flexible to various sensors to be selectively installed and utilized at any time. This includes light and infrared cameras, LiDAR, radar, etc.
\section{Experimental Results}
We conducted two types of experiments. The first group of experiments considered the mini-city as a test bed to evaluate existing algorithms in perception and mapping. The second group of experiments analyzes the proposed smart intersection and evaluates its effectiveness in improving the safety and efficiency of communicating cars at intersections.
\subsection{Extracting the Ground Truth Map of the Mini-city}
A mini-city 2D Ground Truth (GT) map was created on AutoCAD with accurate dimensions of the buildings, roads, and distances between buildings manually measured. This map is compatible with the mapping technique derived from the LiDAR data. Subsequently, the AutoCAD file (DWG file) was used to transform it into a PNG image, which contains the mini-city map. The result is shown in Fig.~\ref{fig:Ground Truth}. The image of the mini-city has a resolution of $1584 \times 1224$ pixels. We use this map as a ground truth to compare other estimated maps. We want to mention the coloring behind the drawing, where black is the borders of the mini-city, grey is a viable area the ARCs can view, and white is driveable roads or buildings.
\subsection{Mapping of the mini-city}
\begin{figure}[h]
    \centering
 \includegraphics[width=1\linewidth]{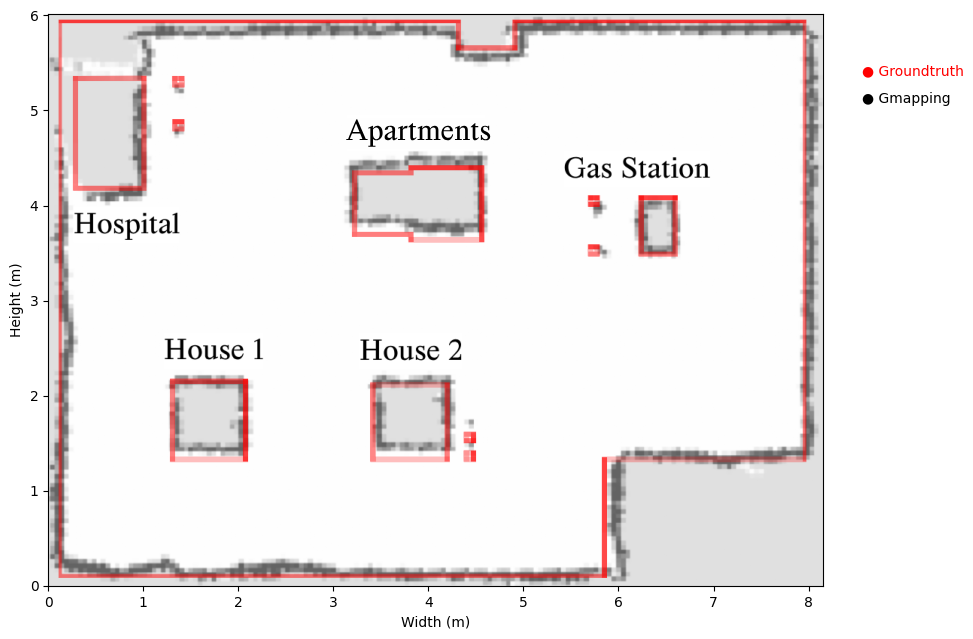}
    \caption{Qualitative comparison of 2D mapping of mini-city using Gmapping (black lines) against Ground Truth (red lines).}
    \label{fig:enter-label}
\end{figure}

\begin{figure}[]
    \centering
    \includegraphics[width=\linewidth]{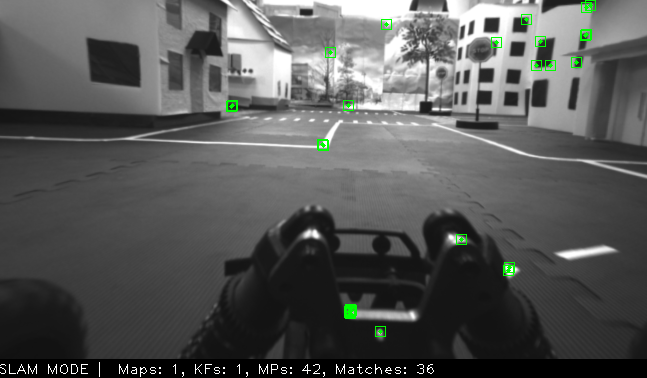}
    \caption{First-person view of mini-city from ARC. Green squares are visual features extracted by ARC's camera and will be used for visual SLAM.}
    \label{fig:FPV}
\end{figure}

Gmapping method~\cite{grisetti2007improved} was chosen to map the mini-city in 2D and can be considered as a baseline for the future mapping algorithms. Gmapping is easy to implement and well documented. We utilize the ARC car’s 2D LiDAR (YDLiDARX4) with an available range of 0.12 to 10 meters and 360 degrees of scanning. The ROS wrapper for OpenSLAM’s Gmapping was utilized to run SLAM on-board. This package provides a laser-based SLAM as a ROS node and allows for the creation of a 2D occupancy grid map.  Gmapping using a Rao-Blackwellized particle filter attempts to solve the independence between localization and mapping. This particle filter maintains a set of particles where each represents a possible path that the robot could have taken, with each particle carrying its version of the map. Odometry data are used to update the weights of the particles. The algorithm then creates a new set of particles, favoring those with higher weights based on the new point cloud collected by LiDAR. All computations were performed onboard utilizing an NVIDIA Jetson Nano (4GB). We found that the best results were obtained by doing at least three loops around the city, when the car is driven at a very low speed. Fig.~\ref{fig:enter-label} shows the estimated map against the GT map created in AutoCAD. We observed that the algorithm recognized most of the walls/obstacles, but came across issues with the corner of the hospital as it is farther from the car. Furthermore, the location of the building was not the most accurate for the Apartments and House 2. The mapping is also evaluated quantitatively in terms of following metrics. Note that the values were converted to meters using the resolution of the map obtained which is about $0.05$ meters/pixel:
\begin{itemize}
    \item \textbf{K-Nearest Neighbor (KNN):}  shows the average distance from each point on one map to its nearest neighbor on the other map. Lower is better.
    \item \textbf{Root Mean Square Error(RMSE):} represents the average difference between corresponding pixel values in the GT and Gmapping maps. The lower is better.
    \item \textbf{Intersection over Union (IoU):} a unitless measure representing the proportion of overlap between the two maps. The range varies from zero to one, higher is better. 
\end{itemize}
The results are summarized in Table~\ref{tab:map_result},  showing  Gmapping provides a map with high accuracy. Moreover, Calculating the KNN distance from GT to Gmapping and vice-versa gives us similar values of KNN Distance (GT to Gmapping). The difference is expected because of the nature of KNN; To elaborate on this difference, let us assume a point A in map 1, with point B as the nearest neighbor in map 2. However, point B's nearest neighbor  may not necessarily be point A. This illustrates the asymmetric nature of KNN. Comparing both maps vice-versa acts as a form of validation. The result shows both values are small and similar, implying a higher level of similarity or alignment between the compared maps, which indicates the map is estimated accurately. In the future, the mini-city can be utilized as a testing bed for exploring more advanced mapping algorithms such as Visual SLAM (VSLAM). This is the case due to its "feature richness" as seen in Fig.~\ref{fig:FPV}. This image shows a great number of distinctive visual features that could be detected and tracked by VSLAM.  
\begin{table} []
    \centering
    \caption{Gmapping evaluation result}
    \begin{tabular}{|l|l|l|l|}
        \hline
        Metric & Value\\
        \hline
        KNN Distance (GT to Gmapping) & 14.15 cm \\
        KNN Distance (Gmapping to GT) & 15.42 cm\\
        RMSE & 0.68 cm\\
        IoU & 0.7415\\
        \hline
    \end{tabular}
    \label{tab:map_result}
\end{table}
\subsection{Depth Estimation in Urban Simulations}
We conducted experiments using the ARC car’s monocular camera, which through these experiments we further look to illustrate the mini-city’s usefulness and capabilities. We analyzed various urban scenarios, to estimate the depth of different usually existing objects in urban areas with respect to the camera (\emph{i.e.,} ego vehicle). The list of objects that has been considered are hospital buildings, trees, cars, pedestrians (humanoid robots), stop signs, as well as highly occluded scenes like a building overlapping another building.

The ARC's RGB camera captured the different scenarios within the mini-city, and from then we extracted the necessary image data from the camera topics in the ROS bags collected from each scenario throughout the experiment. After extracting the required data, different depth estimation algorithms were evaluated and compared: DPT (Dense Prediction Transform) \cite{Ranftl2021}, Depth Anything \cite{depthanything}, and Marigold \cite{ke2023repurposing}.
\begin{figure}
    \centering  \includegraphics[width=1.4\columnwidth,trim={320 0 0 0},clip]{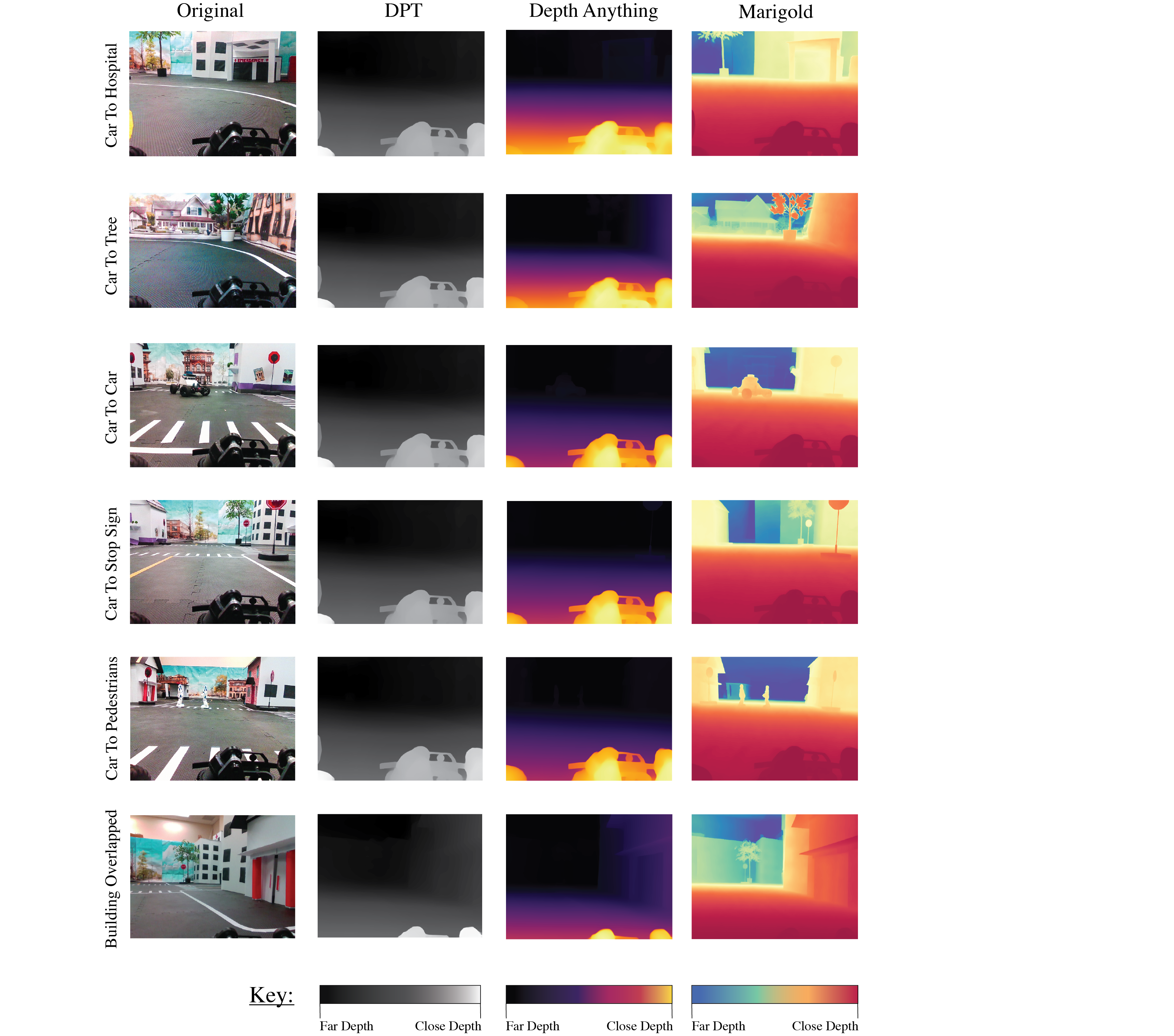}
    \caption{Depth estimation of challenging scenarios in mini-city running different algorithms.}
    \label{fig:depth_experiment}
\end{figure}
\subsection{Algorithm Performance}
Fig.~\ref{fig:depth_experiment} shows the qualitative result of dept estimation algorithms, which indicates DPT and Depth Anything algorithms did not yield significant visual results, aside from obscure shadowing of the expected objects when zoomed in. This likely results from the specific datasets these algorithms were trained on, which did not align well with our mini-city's specific conditions. However, the Marigold algorithm showed more promising results, closely aligning with the different urban scenarios in our mini-city. We also evaluate the baselines quantitatively. The ground truth is obtained by measuring a range of an estimated distance from the car's camera to the intended object manually, when conducting the experiments and then allowed the algorithm to predict the depths of the different scenarios. we focused on three evaluation metrics:
\begin{itemize}
    \item Inference time to gauge the operational efficiency: A lower inference time is desirable as it indicates a faster processing capability.
    \item Mean Absolute Error (MAE) to measure the accuracy of the depth predictions: A lower MAE indicates higher accuracy, as it reflects smaller average errors in depth estimations. 
  \item Mean Relative Error (MRE) to assess the proportionality of the errors in relation to the actual depth values. a lower MRE is preferred as it signifies smaller error proportions relative to the true depth, demonstrating more precise and reliable predictions.
\end{itemize} 
The results highlighted that while DPT and Depth Anything struggled with accuracy and had higher error rates. Marigold performed much better under varied urban conditions, offering lower MAE and significantly reduced MRE; however, it took a significant longer time to process with a high inference time compared to the other two algorithms, which might be more due to limitation of the hardware being used in the actual computation of the algorithm. This underscores the effectiveness of Marigold's training on synthetic and diverse real-world datasets, which enables robust depth estimation even in complex scenarios such as those presented in mini-city. Such findings are crucial for deploying depth estimation technologies in real-world applications where adaptability and accuracy are paramount.
The following paragraph highlights the data resources each baseline is trained on that could explain the difference in their performance performance in our mini-city.
\begin{itemize}
\item \textbf{DPT (Dense Prediction Transformer):}
Trained on a combination of indoor and outdoor scenes, the DPT models utilize data from environments such as NYU Depth V2 ~\cite{Silberman:ECCV12} for indoor settings and KITTI~\cite{Menze2015CVPR} for outdoor scenes. Although they offer broad depth estimation capabilities, they are not tailored for specific scenarios like those presented in mini-city.
\item \textbf{Depth Anything:}Fine-tuned on metric depth information from datasets like NYU Depth V2 and KITTI, Depth Anything is adept at handling both indoor scenarios and driving environments. However, it may not capture the finer details required for the complex urban contexts found in mini-city, such as the specific challenges posed by occlusions and varied object distances.
\item\textbf{Marigold:}
Marigold excels in adapting to a wide range of real-world scenarios, thanks to its initial training on synthetic datasets like Hypersim~\cite{roberts:2021} and Virtual KITTI, which simulate controlled environments. The model is further refined through fine-tuning diverse real-world scenes using advanced latent diffusion techniques, enabling it to effectively handle the unique and varied scenarios encountered in mini-City.
\end{itemize}

\begin{table}[h]
\centering
\caption{Comparison of Depth Estimation algorithms in different scenarios shown in Fig.~\ref{fig:depth_experiment}}
\begin{tabular}{|l|c|c|c|}
 \hline
 Algorithms & Inference Time(s) & MAE(m) & MRE(\%) \\
 \hline
 DPT~\cite{Ranftl2021} & 34.332 & 2.384 & 78.859 \\
 Depth-Anything~\cite{depthanything} & 22.405 & 1.846 &  64.292 \\
 Marigold~\cite{ke2023repurposing} & 603.323 & 0.493 &  13.285 \\
\hline
\end{tabular}
\label{tab:depth_result}
\end{table}

\section{Smart Infrastructure and Connected vehicles}
The smart infrastructure consists of a 3D LiDAR (Velodyne VLP-16) connected to Nvidia Orin processor, and mounted at the southwest corner of the intersection next to the hospital as shown in Fig.~\ref{fig:intersection_scenarios}.

The Velodyne produces the 3D point cloud of the environment and detects and tracks the traffic surroundings of the intersection using depth clustering algorithm \cite{bogoslavskyi17pfg}, where the point clouds are clustered and grouped based on their orientation. This information helps the infrastructure node at the intersection identify any risk of accident that the other cars cannot observe due to their limitation in the field of view. 
We conduct a scenario that includes two cars approaching the smart intersection at the same time. The scenario is shown in Fig.~\ref{fig:intersection_scenarios}.
The car on Road A has the right of way and the car on Road B must yield to the incoming traffic from Road A. These two cars cannot see each other until they are close enough to the intersection. We assume one of the cars is a non-communicating car, non-comm-car in short. Non-comm-car could be a human-driven car without tools to communicate with the infrastructure. However, another car is a communicating car, comm-car in short, and it communicates with the infrastructure. 
 The infrastructure is equipped with 3D LiDAR and it notifies the communicating car if there is any car that is approaching the intersection. Otherwise, the warning is deactivated, which means the communicating car does not stop. 

 This experiment was conducted ten times with two cars approaching the intersection. In the first experiment, the non-communicating car is on road A and communicating car is on road B. The result shows the communicating car stops at an intersection when the other car is approaching, however, it may crash by up to 30 percent due to the inaccuracies of the mapping or delay in communication, leading the car stops too late to avoid the accident. This indicates that inaccuracy in understanding the surroundings may increase the risk of crashes. In the next experiment, the cars locations are swapped such that communicating car is on road A and non-communicating car is on road B. we see that having more accurate mapping at another corner of the intersection improves overall safety as the car stops before the intersection. 
\begin{table}[h]
\centering
\caption{Result of car crashes at the intersection with imperfect data and communication. One car is communicating, one car is not communicating.}
\begin{tabular}{|p{1.75cm}|p{1.75cm}|p{1.65cm}|p{1.5cm}|}
\hline
    comm-car location & no-comm-car location&crashes($\%$)& traveling time (sec)\\ \hline
 Road B & Road A& $30.78\pm 13.05$&$3\pm 0.5$\\ 
 Road A & Road B &$20.82\pm 7.87$ & $2\pm 0.4$\\
\hline
\end{tabular}
    \label{fig:stoppingdistance}
\end{table}
An additional set of experiments was conducted regarding the stopping distance of a communicating car for different approaches toward the intersection and different intersection model parameters. The intersection was approached from all four entry roads (North-, East-, South-, and West-bound) over two different intersection models, where the intersections encompassing the polygon were scaled by $1\times$ and $1.25\times$. The stopping distance was measured as the distance from the rear axle  of the car to the stop line. The results over five trials for each case are summarized in Table~\ref{tab:stoppingdistance}. 
\begin{table}[h]
\caption{Stopping distance (cm) resulting from smart intersection model in a format of $Average\pm std$. $Average$ and $std$ are the averages of the result and the standard deviation respectively. Negative values indicate the car stopped before the stop line, whereas positive values imply the car stopped after the stop line at the intersection.}
\begin{tabular}{|p{0.5cm}|p{1.5cm}|p{1.5cm}|p{1.5cm}|p{1.5cm}|}
\hline
Scale&North-bound &East-bound & South-bound& West-bound
\\
\hline
1.00&$56.4\pm14.6$ & $54.3\pm10.0$  &$-23.7\pm7.9$ & $-18.0\pm2.5$\\
\hline
1.25& $28.3\pm9.6$ &$35.6\pm13.8$&$-48.4\pm4.6$ &$-44.0 \pm1.2$\\ 
\hline
\end{tabular}
\label{tab:stoppingdistance}
\end{table}

These results suggest the intersection model is perhaps off-center towards the South-bound and West-bound sides, resulting in the car's derived "presence" in the intersection being biased in those directions. It also reflects varied standard deviations for certain approach directions, possibly due to varying quality of the localization of the vehicle and to varying qualities of the ground-truth map or localization algorithm used.

\section{Educational Impact}
\begin{figure}[h]
    \centering
\includegraphics[width=0.7\columnwidth,trim={0 0 0 0},clip]{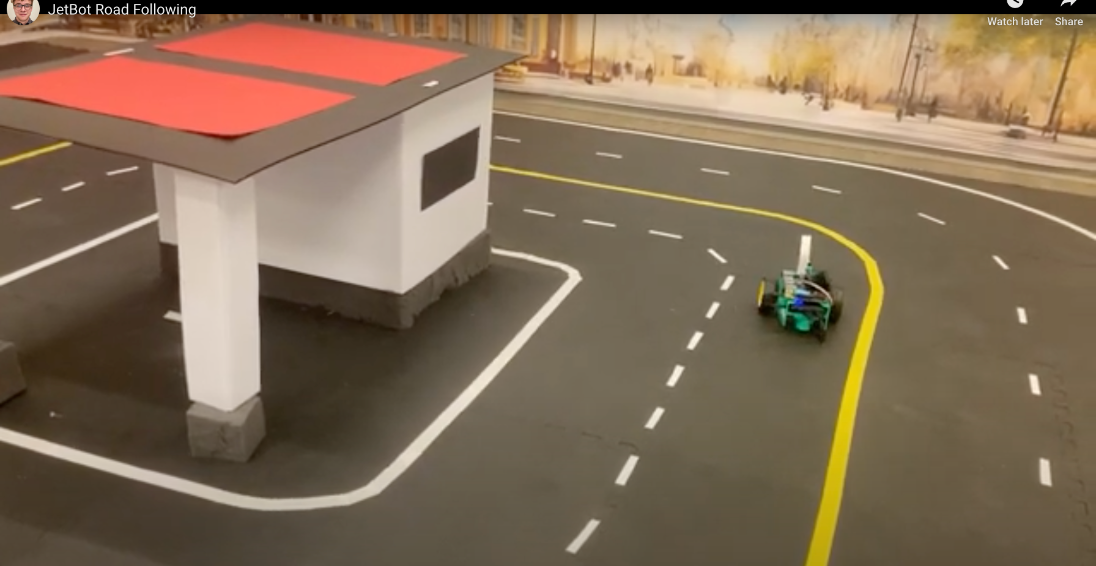}
    \caption{Jetbot following a road in min-icity. This was implemented as the final project for the Artificial Intelligence course at the University of Oklahoma offered in Spring 2024.}
    \label{fig:jetbot}
\end{figure}

The first version of mini-city has been built in October 2023, and the current version was finished in February 2024. Since then,  the mini-city has been used in two courses at the University of Oklahoma; The projects within the coursework involved utilizing the cameras or LiDAR for autonomous navigation, lane detection, obstacle/pedestrian detection, LiDAR navigation, and simultaneous localization and mapping (SLAM).
The course Visual Navigation for Autonomous Vehicles~\cite{carlone2022visual} offered in  Fall 2023, where students utilize robotic cars to visually map the city and autonomously using Kimera~\cite{kimera} and ORB-SLAM3~\cite{orbslam3} navigate in the city. The course Artificial Intelligence offered in Spring 2024, utilized  mini-city as a testbed for students to implement an end-to-end learning method in a Jetbot~\cite{jetbot} to keep it between the lanes in the loop around the gas station in mini-city (Fig.~\ref{fig:jetbot}). In this project the robot follows a road by end-to-end training on on-board camera image data.

\section{Conclusion}
In this paper, we present our smart mini-City as a pivotal development for bridging the gap between simulated environments and real-world applications in the field of autonomous connected vehicles and intelligent transportation systems. Through our experiments, we have validated the mini-city as a crucial and advancing test bed for smart city technology and evaluating autonomous vehicle systems as a whole, which we demonstrated by conducting our mapping experiment as well as depth estimation the mini-city among different algorithms through a variety of challenging scenarios. Our work with the smart infrastructure at our intersection also further demonstrates enhancements in traffic management and safety, as well as setting the stage for future research into more intelligent transportation systems. Looking ahead, the mini-city is poised to expand its role, incorporating more complex scenarios and interactive elements like responsive traffic signals and pedestrian crossing at interactions. Finally, as the mini-city contributes to the development of urban mobility research in different simulated environments, it naturally fosters interdisciplinary research that converges urban planning, civil engineering, and computer science, while also maintaining cost-effectiveness so any laboratory can construct one and produce a similar testing bed. As we enhance and evolve the mini-city, it is set to contribute to shaping the development of smarter, safer, and more efficient urban environments.
\section*{ACKNOWLEDGMENT}
We would like to thank the School of Computer Science at the University of Oklahoma for supporting our efforts. This project is funded by the Data Institute for Societal Challenges (DISC) at the University of Oklahoma.
\bibliographystyle{IEEEtran}

\bibliography{main.bib}
\balance

\end{document}